\documentclass[lettersize,journal]{IEEEtran}
\pdfoutput=1
\usepackage{amsmath,amsfonts}
\usepackage{algorithmic}
\usepackage{array}
\usepackage[caption=false,font=normalsize,labelfont=sf,textfont=sf]{subfig}
\usepackage{textcomp}
\usepackage{stfloats}
\usepackage{url}
\usepackage{verbatim}
\usepackage{graphicx}
\usepackage{bm}
\usepackage{pifont}
\usepackage{color}
\usepackage{multirow}
\usepackage{booktabs}
\usepackage{balance}
\usepackage{hyperref}
\def\BibTeX{{\rm B\kern-.05em{\sc i\kern-.025em b}\kern-.08em
		T\kern-.1667em\lower.7ex\hbox{E}\kern-.125emX}}
\begin{document}
\title{VTPNet for 3D deep learning on point cloud}
\author{Wei Zhou*†, Weiwei Jin*, Qian Wang, Yifan Wang, Dekui Wang, Xingxing Hao, Yongxiang Yu
\thanks{Manuscript created April, 2023; 
*Wei Zhou and Weiwei Jin contributed equally in this paper.
†Corresponding author: Wei Zhou.

Wei Zhou, Weiwei Jin, Qian Wang, Yifan Wang, Dekui Wang, Xingxing Hao are with the School of Information Science and Technology, Northwest University, Xi'an 710127, China (e-mail: mczhouwei12@gmail.com; weiweijin1109@gmail.com
)  

Yongxiang Yu is with the Department of Electrical and Automatic Engineering, East China Jiaotong University, Nanchang 330013, China
}}
	
	\markboth{Journal of \LaTeX\ Class Files,~Vol.~18, No.~9, September~2020}%
	{How to Use the IEEEtran \LaTeX \ Templates}
	
	\maketitle
        \begin{abstract}
        Recently, Transformer-based methods for point cloud learning have achieved good results on various point cloud learning benchmarks. However, since the attention mechanism needs to generate three feature vectors of \emph{query}, \emph{key}, and \emph{value} to calculate attention features, most of the existing Transformer-based point cloud learning methods usually consume a large amount of computational time and memory resources when calculating global attention. To address this problem, we propose a Voxel-Transformer-Point (VTP) Block for extracting local and global features of point clouds. VTP combines the advantages of voxel-based, point-based and Transformer-based methods, which consists of Voxel-Based Branch (V branch), Point-Based Transformer Branch (PT branch) and Point-Based Branch (P branch). The V branch extracts the coarse-grained features of the point cloud through low voxel resolution; the PT branch obtains the fine-grained features of the point cloud by calculating the self-attention in the local neighborhood and the inter-neighborhood cross-attention; the P branch uses a simplified MLP network to generate the global location information of the point cloud. In addition, to enrich the local features of point clouds at different scales, we set the voxel scale in the V branch and the neighborhood sphere scale in the PT branch to one large and one small (large voxel scale \& small neighborhood sphere scale or small voxel scale \& large neighborhood sphere scale). Finally, we use VTP as the feature extraction network to construct a VTPNet for point cloud learning, and performs shape classification, part segmentation, and semantic segmentation tasks on the ModelNet40, ShapeNet Part, and S3DIS datasets. The experimental results indicate that VTPNet has good performance in 3D point cloud learning. The source code and pre-trained model of VTPNet will be released on \url{https://github.com/aaaajinweiwei/VTPNet}.
        \end{abstract}
	
	\begin{IEEEkeywords}
		Point cloud, deep learning, transformer, classification, segmentation.
	\end{IEEEkeywords}

	\section{Introduction}
	\label{sec:introduction}
	\IEEEPARstart{W}{ith} the rapid development of technologies such as LiDAR, the acquisition of 3D point cloud data is becoming increasingly convenient. Point cloud learning has become increasingly important in areas of computer vision such as robotics, autonomous driving, virtual reality, and augmented reality. However, due to the irregularity of point clouds, how to effectively capture semantic information from point cloud is still a problem.

        \begin{figure}[ht]
		\centering
		\includegraphics[width=0.45\textwidth]{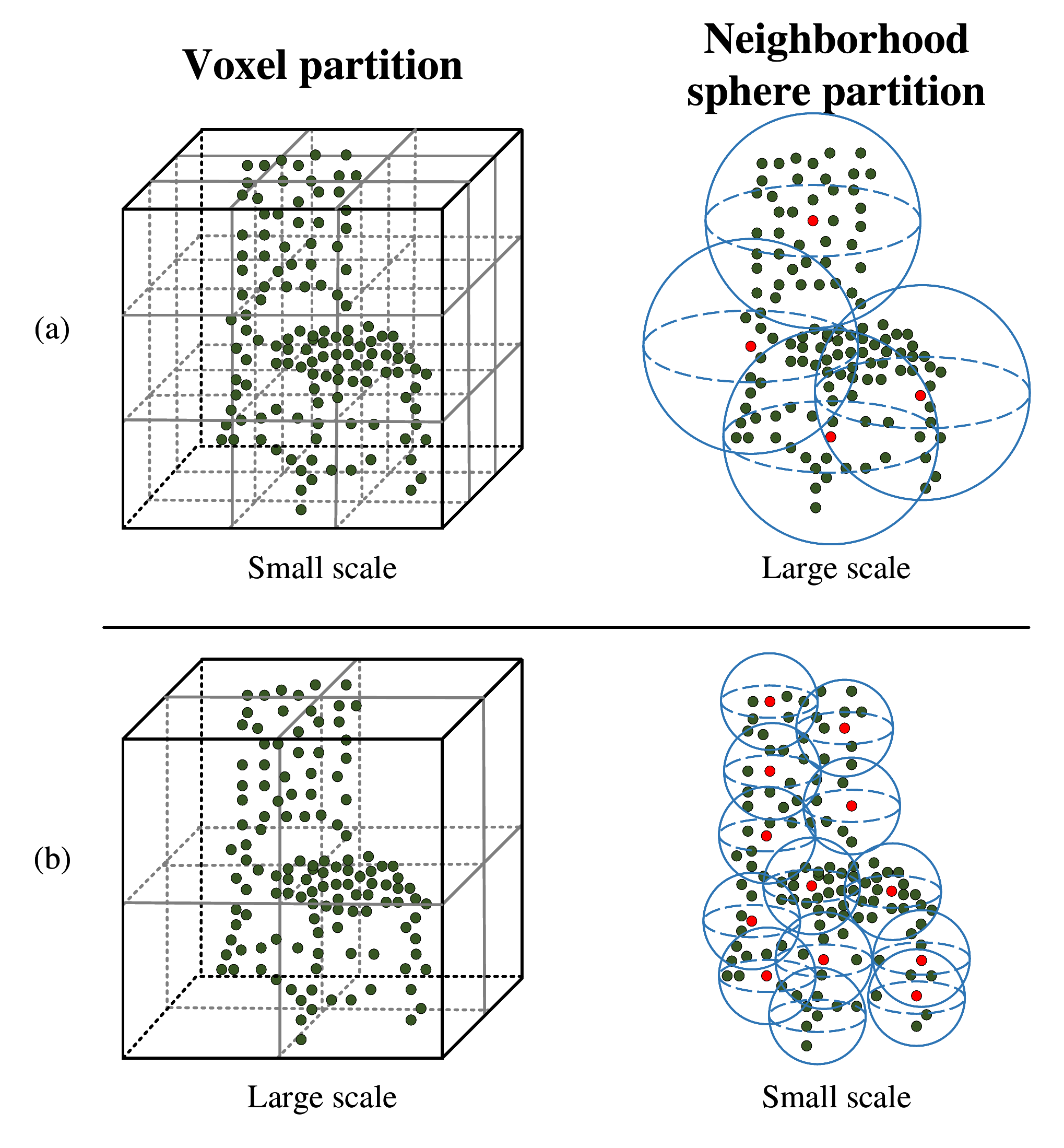}
		\caption{Strategy ``One large and one small" enriches local features of point cloud at different scales. The top half (a) of the figure indicates that when the voxel scale is small, the neighborhood sphere scale is set to be large. The bottom half (b) indicates that the neighborhood sphere scale is set small when the voxel scale is large.} \label{fig:scale_set}
	\end{figure}
	
	There are various methods for processing irregular point cloud, among which voxel-based \cite{octnet,o-cnn,pointgrid,ben20173d,voxelnet} and point-based \cite{pointnet,pointnet++,dgcnn,so-net,pointweb,pointconv,3d-gcn} methods are widely used. The voxel-based methods divide point clouds into regular voxel grids, and then perform feature learning on the regular voxels. This type of method owns high computational efficiency, but it will inevitably lose some detail information, and high voxel resolution would sacrifice more GPU memory and computational efficiency, thus limiting the ability of the model. In contrast, the point-based methods can retain the accuracy of point location with much lower occupancy of GPU memory than voxel-based methods, but processing this type of method would bring random memory access, thus would result in inefficiency of model.
	
	Transformer \cite{attention} has demonstrated its excellent performance in the field of natural language processing (NLP). Recently, Transformer is extensively applied in the area of computer vision for 3D point cloud learning, and has achieved good performances \cite{pt,pct,pvt,pointr}. The core of Transformer is the modeling of self-attention mechanism. The self-attention mechanism passes the input features through three trainable linear layers to obtain three feature vectors of $query$, $key$, and $value$ respectively, and obtains the attention weight by calculating the dot product of the $query$ and $key$ vectors, the self-attention encoding is the weighted summation of attention weights and $value$ vectors. However, there are still some problems in the existing Transformer-based methods: most of them extract global features through Transformer, and three feature vectors of $query$, $key$, and $value$ need to be generated during the attention calculation process. However, point cloud data are often in large size, which consumes a lot of time and memory resources in the process of extracting global features, Moreover, a simple global attention mechanism cannot effectively extract features of point clouds.

	In this paper, we learn to research how to design an effective Transformer-based learning network for 3D point cloud while avoiding the weakness of previous point cloud Transformer-based methods, and proposes a new deep learning network for 3D point cloud processing named VTPNet which integrates the advantages of the voxel-based and the point-based methods. Learning that the voxel-based methods can obtain effective coarse-grained local point cloud features with low time consumption, while the point-based methods can extract fine-grained local point cloud features, and applying Transformer in local neighborhoods can greatly reduce the cost of time efficiency and GPU memory, therefore we combine the voxel-based and the point-based Transform branches to extract feature information of point cloud, thus to obtain two different types of local features. In the voxel-based branch, we conduct 3D convolution on voxel features to obtain voxel-level local features, while in the point-based Transform branch, we adopt the self-attention mechanisms within the neighborhood of points to obtain local features, and then calculate cross-attention between neighborhoods to obtain point-level local features. In addition, we set the voxel scale in the voxel-based branch and the neighborhood sphere scale in the point-based Transform branch, one of which is large, and the other is small, as shown in Fig.\ref{fig:scale_set}, thus to obtain local features at different scales, thereby further improving the feature description ability of the model. However, the outputting of the voxel-based and point-based Transform branches only contain single local features, which lack global location information, so we add a simple point-based branch to extract the global location information of the point cloud, thus to supplement the other two branches to obtain the depth features of the point cloud. These three branches mentioned above constitute the feature extraction block called VTP (Voxel, Transform and Point) which holds an important component of our proposed VTPNet backbone network.
	
	The VTP block proposed in this paper can be applied to various 3D point cloud classification and segmentation tasks as a feature extraction network. We use VTP as the core part to construct 3D semantic segmentation backbone network named VTPNet.
	
	The main contributions of this paper are summarized as follows:
	
	\begin{itemize}
		\item We propose a feature extraction module VTP which integrates voxel-based, Transformer-based and point-based methods.
		\item We design two branches for extracting local features of point clouds with different grained sizes. The voxel-based branch is used to extract coarse-grained local features of point clouds, and the point-based Transform branch is used to extract fine-grained local features of point clouds.
		\item We set the voxel scale in the voxel-based branch and the neighborhood sphere scale in the point-based Transform branch, one of which is large, and the other is small, thus to obtain local features at different scales (large voxel scale \& small neighborhood sphere scale or small voxel scale \& large neighborhood sphere scale).
		\item We design a 3D semantic segmentation backbone network VTPNet based on VTP for different datasets, and conducts experiments on ShapeNet Part \cite{shapenetpart}, S3DIS \cite{s3dis} and ModelNet40 \cite{modelnet40} datasets, with experimental results of 85.8\%, 66.3\% and 93.2\% respectively, which demonstrates the effectiveness of our proposed network for 3D point cloud learning.
	\end{itemize}  
	
	\section{Related Work}
	\label{sec:related_work}
		\subsection{Voxel-based and Multiview-based methods}
		The excellent performance of convolutional neural networks (CNN) in image processing has led to its application in 3D point cloud processing. However, typical CNN cannot be directly applied to irregular point cloud data, so many researchers firstly convert point cloud data into regular data structures before performing CNN, this type of method includes the voxel-based and the multiview-based methods. The voxel-based methods firstly convert point cloud into regular voxel grids, and then use 3D convolution for processing \cite{modelnet40,vv-net,voxnet,roynard2018classification}. However, information loss is inevitable during the voxelization for point cloud, especially in low voxel resolution. The typical way to improve the performance of the voxel-based model is usually to increase the voxel resolution, but high voxel resolutions will consume a large amount of GPU memory, and 3D CNN on them also requires a high time cost. The multiview-based methods typically map 3D shapes onto the two-dimensional plane of multiple views, and use the mapped views as the input of the model, it has achieved good performances in 3D shape classification \cite{3d-r2n2,su2015multi,rotationnet,qi2016volumetric,tatarchenko2018tangent,atzmon2018point}. However, it’s difficult to extend the multiview-based methods to 3D point cloud tasks such as semantic segmentation and point cloud completion.
		
\subsection{Point-based methods}
Qi et al proposed PointNet  by using multiple shared multi-layer perceptrons (MLPs) to extract point cloud features, this is the first deep neural network that directly uses 3D point cloud as input, and achieved good performances \cite{pointnet}. 
After PointNet, Qi et al. further proposed PointNet++ to obtain different levels of point cloud features \cite{pointnet++}, which is inspired by the U-Net structures \cite{cciccek20163d} and still adopts MLPs to extract features. 
Xu et al. conducted MLP over the fused features of Euclidean and eigenvalue spaces to extract the point cloud features \cite{xu2020geometry}. 
Hu et al. proposed RandLANet that directly processes point clouds in large scenes \cite{hu2020randla}. 
Rather than the innovation in model architecture, PointNeXt improves on the basis of PointNet++ by adding an InvResMLP module behind the set abstraction layer and adopting an improved training strategy~\cite{pointnext}.
Zhang et al. proposed I2P-MAE by ussing self-supervised pre-training to learn 2D knowledge by multi-view features, thus to guide 3D MAE \cite{zhang2022learning}. 
Yin et al. proposed DCNet to address the problem of information loss caused by random downsampling \cite{yin2023dcnet}.
Li et al. proposed a fast data structuring to solve the problem of high time cost of existing sampling and grouping methods \cite{li2022psnet}.
		
Subsequent works consider redefining the convolution for irregular point cloud data. Li et al. proposed PointCNN by trained an X-transformation to regularize irregular point cloud data, and then used convolution for the regularized data \cite{li2018pointcnn}. Wu et al. considered that the point features of point cloud are also related to its K-nearest neighbors, and the weights are different for different locations, therefore, they use MLP to learn the weight function within the neighborhood \cite{pointconv}. Wang et al. proposed DGCNN by designing the EdgeConv operator, iterating repeatedly and dynamically calculating the local features of the point cloud, thus to learn the local geometric information of the point cloud \cite{dgcnn}. Zhao et al. believed that the points within a neighborhood are related to each other, so it learns the relationships between points neighboring within a neighborhood, so as to enhance the local features of point cloud \cite{pointweb}. Thomas et al. proposed a convolution method called KPConv that can be used directly on point clouds without any intermediate representation \cite{thomas2019kpconv}. Wiersma et al. constructed an anisotropic convolutional layer that combines geometric operators from vector operations \cite{wiersma2022deltaconv}.
		
In addition, some researchers have fused point-based and voxel-based methods, this type of model combines the advantages of these two methods, it owns the ability to extract deep features while consuming low time and GPU memory. The PVCNN proposed by Liu et al. uses a point-based method to obtain the features of each point, while using voxel-based method to obtain the local features of the point cloud, and fuses these two features as the extracted point cloud features \cite{liu2019point}. Shi et al. constructed a voxel set abstraction layer, which concatenates the aggregated keypoint features of the point cloud and the corresponding aggregated voxel features as the point cloud features \cite{shi2020pv}. Ye et al. proposed SPVFE and SVPFE, which iterating voxel features and point features continuously, thus to make the features propagate across different modules \cite{ye2021drinet}.
		
		\subsection{Transformer in Point Cloud}
		In recent years, Transformer has achieved great success in the field of NLP, which has inspired the development of 3D point cloud processing \cite{attention,wu2019pay,devlin2018bert,dai2019transformer,yang2019xlnet}. There have been many previous efforts to extract point cloud features by using Transformer. Point Transformer \cite{pt} proposed by Zhao et al. introduced an attention operator based on subtraction to capture local information of point clouds. Zhang et al. proposed Patchformer \cite{patchformer} which introduces Patch Attention to learn the global shape information of point clouds with a lower cost of time, and also proposed a lightweight multi-scale attention module to generate the multi-scale features for the model. Point Cloud Transformer \cite{pct} proposed by Guo et al. constructs an offset-attention module to encode point cloud features and extract local features by using an attention module between point sets. Lai et al. proposed Stratified Transformer \cite{Lai2022StratifiedTF} which uses a new sampling strategy for key to expand the acceptance domain of the model and reduce computational costs. Zhang et al. proposed Spark Window Attention (SWA) module to captures coarse-grained local features from voxels, and uses two different attention variants to extract global fine-grained features at different scales \cite{pvt}. Voxel Set Transformer \cite{he2022voxel} proposed by He et al. introduces Set-Attention to solve the problem of large time consumption for global Transformers. Huang et al. \cite{huang2023lcpformer} proposed a Local Context Propagation (LCP) module that utilizes the overlapping parts of adjacent local regions to transmit local structural information. Song et al. \cite{song2022lslpct} proposed a new Transformer framework to enhance Transformer's ability to learn local semantic features, and designed an efficient self-attention mechanism to capture finer-grained local semantic features.

	\begin{figure*}
		\centering
		\includegraphics[width=0.8\textwidth]{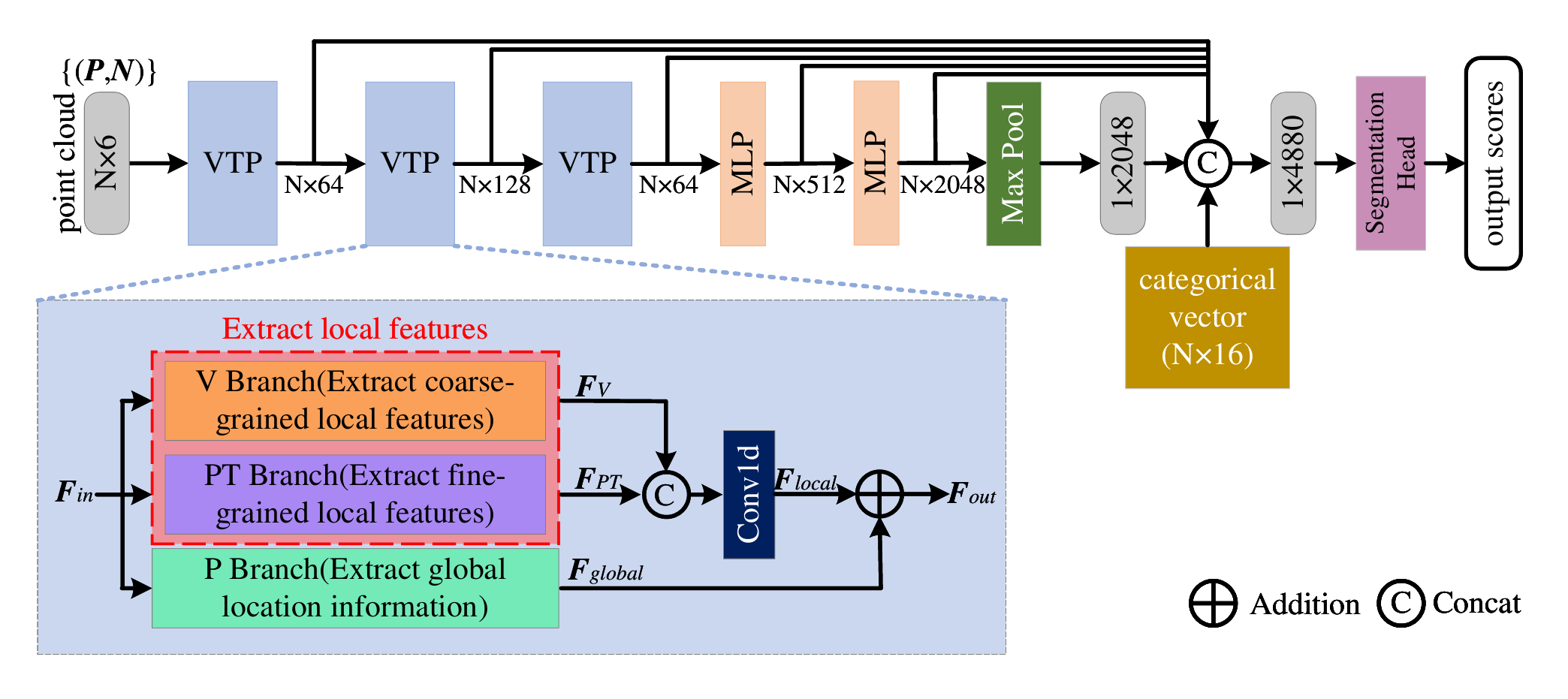}
		\caption{VTPNet deep learning network (ShapeNet Part). VTP enriches the coarse-grained and fine-grained features of point clouds. Its backbone consists of three VTPs and two MLPs. VTP is mainly composed of V branch, PT branch and P branch. V branch and PT branch (dotted red box) are used to extract local coarse-grained and fine-grained features of the point cloud respectively, and P branch provides global location information of the point cloud. MLP consists of one-dimensional convolution, batch normalization and nonlinear activation function. The numbers below the arrow indicate the feature dimension of the transmission.} \label{fig:vtpnet}
	\end{figure*}

	\section{Method}
	\label{sec:method}
	Due to the recent excellent performance of Transformer in the field of computer vision, we have also tried to apply it to the semantic segmentation for 3D point clouds. Our network for part segmentation is shown in Fig.\ref{fig:vtpnet}, its backbone consists of VTP blocks and MLPs with different scales. The input for our network is the point cloud with $N$ points $\left\{(\bm{P},\bm{N})\right\}=\{(p_{i},n_{i})|i=1,2,...~N\}$ which includes the coordinates $\bm{P}=\{p_{i}\in\mathbb{R}^{3}|i=1,2,\ldots N\}$ and the normal $\bm{N}=\{n_{i}\in\mathbb{R}^{3}|i=1,2,\ldots N\}$. We concatenate the global features obtained through maximum pooling, the features generated by the previous blocks and label vectors, and then send them to the segmentation head to obtain the output scores.
	
	As shown in the lower left part of Fig.\ref{fig:vtpnet}, VTP takes $\bm{F}_{in}\in\mathbb{R}^{N\times C_{in}}$ as input and $\bm{F}_{out}\in\mathbb{R}^{N\times C_{out}}$ as output, where $N$ is the number of points in the point cloud, $C_{in}$ is the number of input feature channels, and $C_{out}$ is the number of output feature channels. Our VTP block consists of three parts: Voxel-Based Branch (V Branch), Point-Based Transformer Branch (PT Branch), and Point-Based Branch (P Branch). In VTP block, both V Branch and PT Branch are used to extract local information from point clouds, where V Branch is used to extract voxel-level local features $\bm{F}_{V}\in\mathbb{R}^{N\times C_{o u t}}$ (coarse-grained local features), and PT Branch is used to extract point-level local features $\bm{F}_{PT}\in\mathbb{R}^{N\times C_{o u t}}$ (fine-grained local features), we concatenate the output features of these two branches as the local feature representation of points. P Branch extracts the global location information $\bm{F}_{global}\in\mathbb{R}^{N\times C_{o u t}}$ of points, thereby enriching the feature representation of points. In addition, to enrich the local features of point clouds with different scales, we reduce the neighborhood sphere radius in the PT Branch (while increasing the number of key points) when the scale of the voxel block in the V Branch is large; Similarly, when the scale of the voxel block in the V Branch is small, we increase the neighborhood sphere radius in the PT Branch (while reducing the number of key points). We will demonstrate the effectiveness of this idea in Section \ref{sec:exp_ablation}.
	
		\begin{figure}
			\centering
			\includegraphics[width=0.49\textwidth]{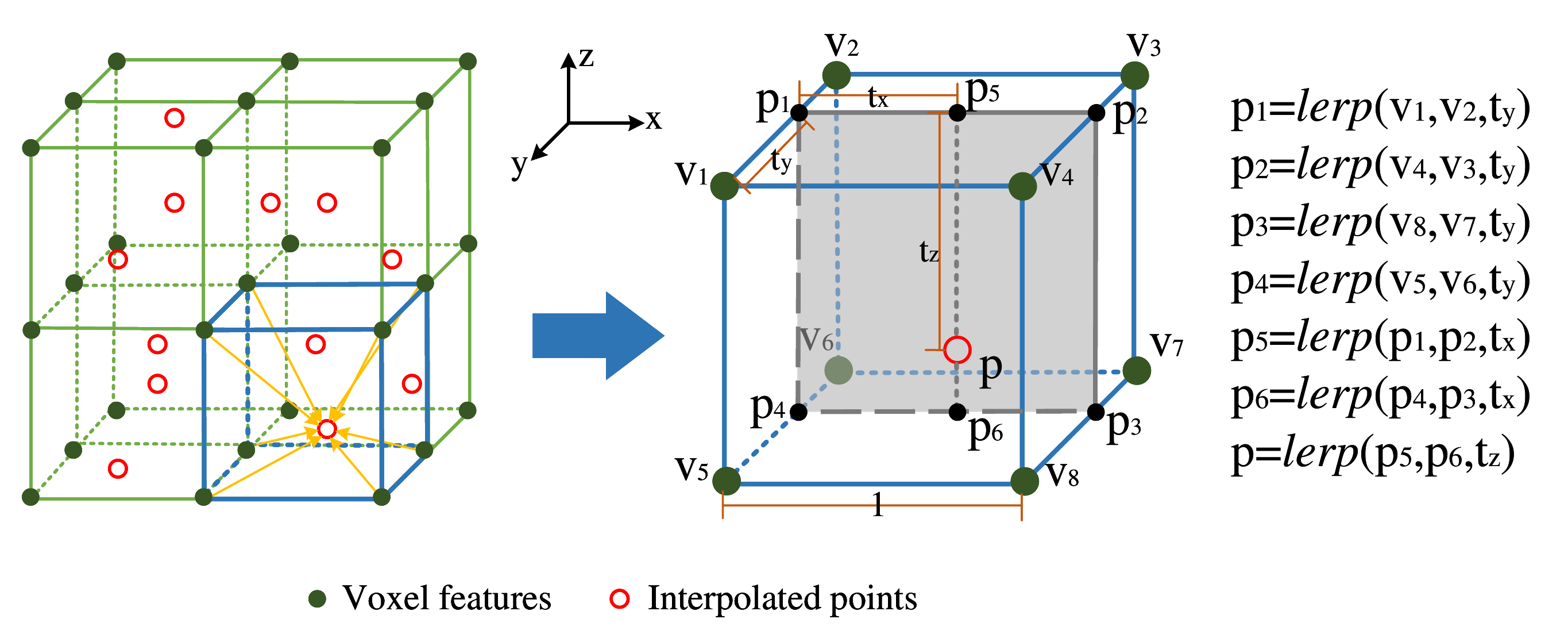}
			\caption{Trilinear interpolation. $lerp(a,b,t)=a\cdot(1-t)+b\cdot t$ denotes the linear interpolation between $a$ and $b$ with proportion of $t$.} \label{fig:tril_inter}
		\end{figure}

		\subsection{Voxel-Based Branch}
		To effectively obtain local information of point clouds with less costs of GPU memory and time efficiency, we use the voxel-based method to capture the voxel-level local features of point clouds. This branch mainly includes three parts: voxelization, feature aggregation and interpolation. The voxelization divides the point clouds into regular voxel grids, then adopts a series of 3D CNNs to aggregate the voxel features. Finally, the voxel features are interpolated into the common domain of point cloud (see Fig.\ref{fig:tril_inter}), thus we obtain the voxel-level local features $\bm{F}_V$, the detail can refer to \cite{liu2019point}.
		
		\begin{figure*}
			\centering
			\includegraphics[width=1\textwidth]{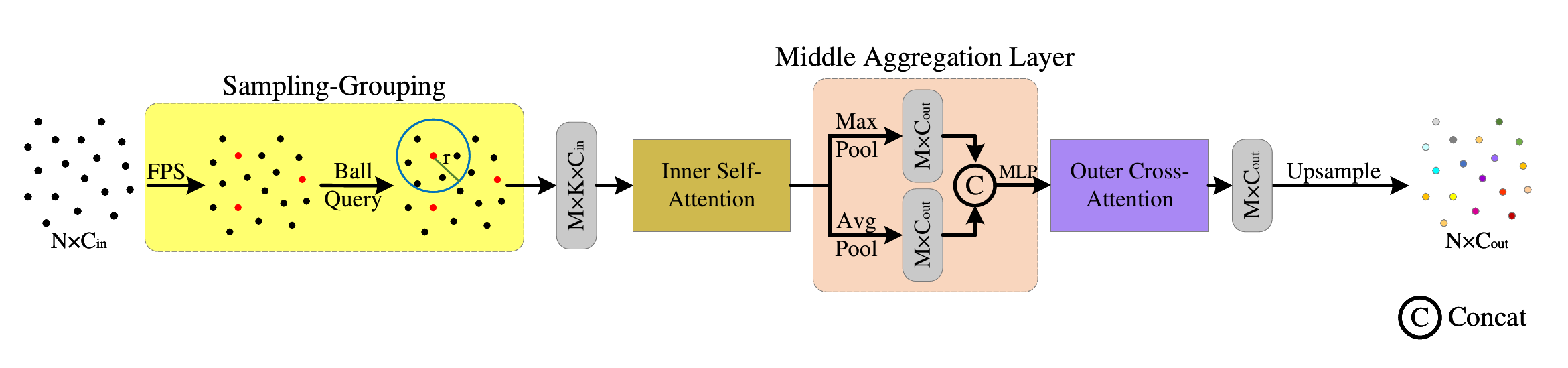}
			\caption{Point-Based Transformer Branch. The PT branch consists of five parts: \ding{172} Sampling-Grouping uses FPS to sample keypoints and divides local-overlapping neighborhoods with the keypoints as the center and $r$ as the radius. \ding{173} Inner Self-Attention calculates self-attention for points in the local neighborhood to obtain deep point cloud features. \ding{174} The Middle Aggregation Layer performs max pooling and average pooling on the deep point cloud features obtained from Inner Self-Attention, then we concatenate these two pooling features and process them with MLP which is composed of linear layer, batch normalization and activation function. \ding{175} Outer Cross-Attention computes cross-attention for keypoints. \ding{176} Finally, the keypoints are upsampled to the original point cloud size by using the nearest neighbor interpolation.} \label{fig:pt_branch}
		\end{figure*}
		
		\subsection{Point-Based Transformer Branch}
		The previous V Branch only extracts shallow coarse-grained features of the entire voxels, without extracting the deep fine-grained features of interior points within the voxel. In order to enrich the local feature representation of point clouds, we adopt the PT branch to extract the point-level local features of the point clouds: Firstly, we use the self-attention mechanism to update the features of the neighboring points within the local neighborhood of keypoints (Inner Self-Attention); Next we obtain the features of the keypoints through the middle aggregation layer; Then we perform cross-attention on all keypoints (Outer Cross-Attention); Finally, we upsample the features of keypoints to the original point cloud (Feature Interpolation).
		
		\textbf{Sampling-Grouping.} As shown in the left yellow module of Fig.\ref{fig:pt_branch}, we firstly divide the point cloud into multiple neighborhoods, thus to extract the local features of the point cloud. Most of the methods typically adopts KNN or ball query to divide the neighborhoods, the KNN method can fix the number of points in the neighborhood, while the ball query method could make the neighborhood have a fixed scale which uses a neighborhood sphere with a radius of $r$ to divide the neighborhood, which is more suitable for extracting local features. In this paper, we use the ball query to divide the local domain (the neighbor points $K$ will be fixed in our method). Before using the ball query to partition the neighborhood spheres, we downsample the point cloud to obtain the keypoint, and set it as the center of the neighborhood sphere. To make the keypoints more evenly distributed in the point cloud, we use the furthest point sampling method (FPS) to downsampling the point cloud $\bm{P}=\{p_{1},p_{2},\ldots,p_{N}\}$, thus we obtain keypoints $\bm{P}_{k}=\{p_{k_{1}},p_{k_{2}},\ldots,p_{k_{M}}\}\subset \bm{P}$, where $N$ is the number of points in point cloud and $M$ is the number of keypoints. The corresponding features of keypoints are denoted as $\bm{F}_{k}=\left\{f_{k_{i}}\in\mathbb{R}^{1\times C_{i n}}|i=1,2,\ldots,M\right\}$.
		
		\begin{figure}
			\centering
			\includegraphics[width=0.5\textwidth]{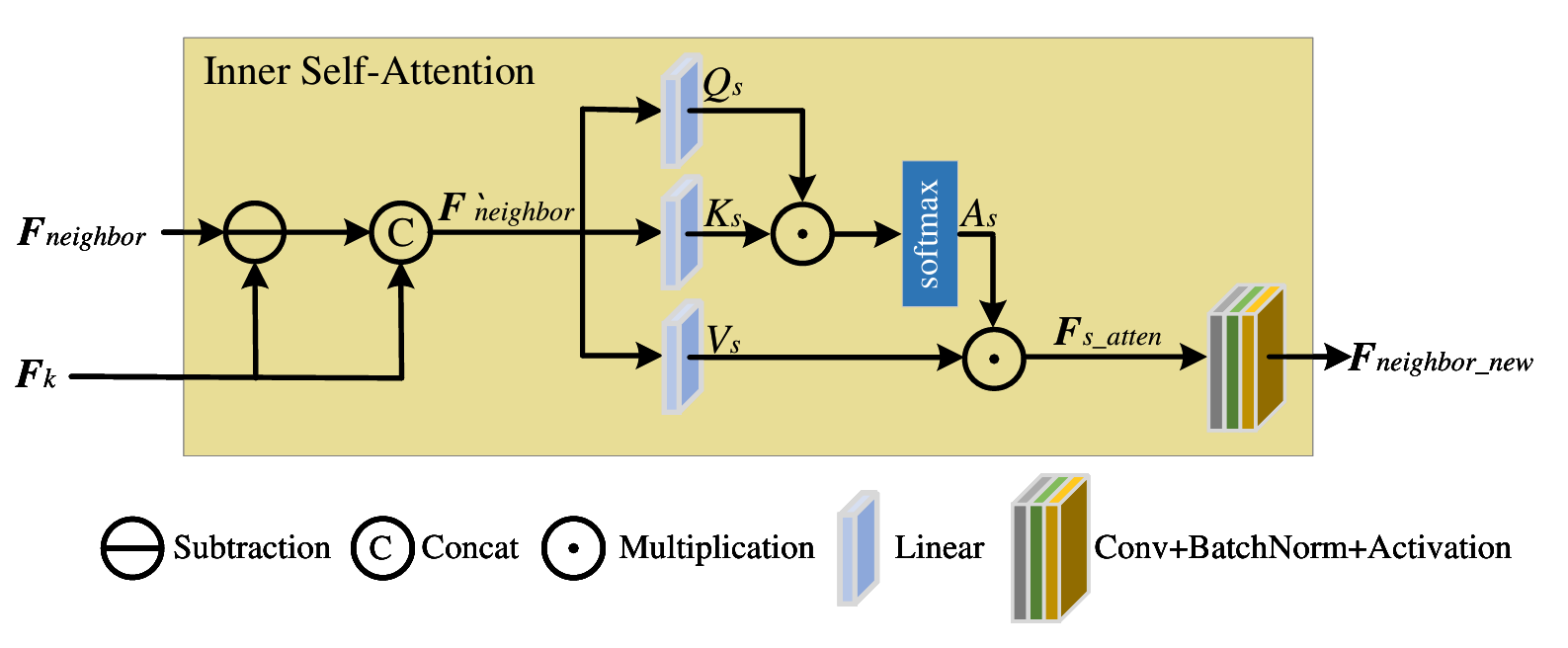}
			\caption{Inner Self-Attention. This module concatenates the features of keypoints and the difference features between neighbor points and keypoints, then calculates the self-attention of the features obtained by concatenation.} \label{fig:inner_atten}
		\end{figure}
	
		\textbf{Inner Self-Attention.} As shown in Fig.\ref{fig:inner_atten}, in PT Branch, we use the traditional self-attention mechanisms for local feature extraction. Assuming $\bm{F}_{neighbor_{i}}\in\mathbb{R}^{K\times C_{in}}$ are the features of the points in the neighborhood of the $i$th key point $p_{k_i}$, where $C_{in}$ represents the feature channels of the point, and $K$ represents the number of neighboring points. To obtain local features of point clouds more effectively, we use ${\bm{F}}'_{neighbor_{i}}\in\mathbb{R}^{K\times2C_{in}}$ instead of $\bm{F}_{neighbor_{i}}$ for self-attention calculation:
		\begin{equation}
			\bm{F}_{neighbor_{i}}^{\prime}=concat(\bm{F}_{neighbor_{i}}-f_{k_{i}},f_{k_{i}}) \label{eq:1}
		\end{equation}
		where $f_{k_i}$ is the feature of the keypoint $p_{k_i}$, and $concat$ represents the feature concatenation operation. Then the $query$ ($Q_s$), $key$ ($K_s$) and $value$ ($V_s$) matrices used for attention calculation can be expressed as:
		\begin{equation}
			\begin{cases}
				Q_{s}={\bm{F}}'_{neighbor_{i}}\cdot W_{q\_s},Q_{s}\in\mathbb{R}^{K\times C_{out}} \\
				K_{s}={\bm{F}}'_{neighbor_{i}}\cdot W_{k\_s},K_{s}\in\mathbb{R}^{K\times C_{out}} \\
				V_{s}={\bm{F}}'_{neighbor_{i}}\cdot W_{v\_s},V_{s}\in\mathbb{R}^{K\times C_{out}} \label{eq:2}
			\end{cases}
		\end{equation}
		where $W_{q\_s}$, $W_{k\_s}$ and $W_{v\_s}$ are the shared and learnable linear transformations matrices, and $C_{out}$ is the feature channels of the $query$, $key$, and $value$ vectors. Next we calculate the attention weight $A_s$, and multiply the attention weight $A_s$ by the $value$ matrix, thus to obtain the attention encoding $\bm{F}_{s\_atten}$:
		\begin{align}
			&A_{s}=softmax(Q_{s}\cdot K_{s}^{\ T}),A_{s}\in\mathbb{R}^{\mathbb{K}\times K_{l}}) \label{eq:3} \\
			&\bm{F}_{s\_atten}=A_{s}\cdot V_{s},\bm{F}_{s\_atten}\in\mathbb{R}^{K\times C_{out}} \label{eq:4}
		\end{align}
		where $softmax$ represents the normalization function, which controls the attention weight in the range of $[0,1]$.
		
		To aggregate the features of points within the neighborhood, we apply convolution to the attention coding $\bm{F}_{s\_atten}$ and then undergo a batch normalization and a nonlinear activation function to obtain the $i$th updated neighborhood features $\bm{F}_{neighbor\_new_{i}}\in\mathbb{R}^{K\times C_{out}}$:
		\begin{equation}
			\bm{F}_{neighbor\_new_{i}}=\alpha(\beta(\omega(\bm{F}_{s\_atten}))) \label{eq:5}
		\end{equation}
		where $\omega$ represents the convolution operation, $\beta$ is the batch normalization function, and $\alpha$ is the nonlinear activation function. Then we pass the updated neighborhood features $\bm{F}_{neighbor\_new_i}$ through the Middle Aggregation Layer to obtain key point features. 
		
		\textbf{Middle Aggregation Layer.} As shown in the pink module of Fig.\ref{fig:pt_branch}, we perform max pooling and average pooling on the $i$th updated neighborhood features $\bm{F}_{neighbor\_new_i}$, thus we obtain $f_{max_{i}}\in\mathbb{R}^{1\times C_{out}}$ and $f_{mean_{i}}\in\mathbb{R}^{1\times C_{out} }$ features respectively:
		\begin{equation}
			\begin{cases}
				f_{max_{i}}=max(\bm{F}_{neighbor\_new_i}) \\
				f_{mean_i}=mean(\bm{F}_{neighbor\_new_i}) \label{eq:6}
			\end{cases}
		\end{equation}
		where $max$ and $mean$ represent maximum pooling and average pooling operations, respectively. Then we process the concatenate features of $f_{max_i}$ and $f_{mean_i}$ with MLP to get the updated $i$th keypoint feature:
		\begin{equation}
			f_{k\_new_{i}}=\zeta(concat(f_{max_{i}},f_{mean_{i}})),f_{k\_new_{i}}\in\mathbb{R}^{1\times C_{out}} \label{eq:7}
		\end{equation}
		where $\zeta$ denotes an MLP composed of linear layers, batch normalization, and activation functions. Then all the updated keypoint features are presented as:
		\begin{equation}
			\bm{F}_{k\_new}=\left\{f_{k\_new_{i}}\in\mathbb{R}^{1\times C_{out}}|i=1,2,...,M\right\} \label{eq:8}
		\end{equation}
		
		\begin{figure}
			\centering
			\includegraphics[width=0.5\textwidth]{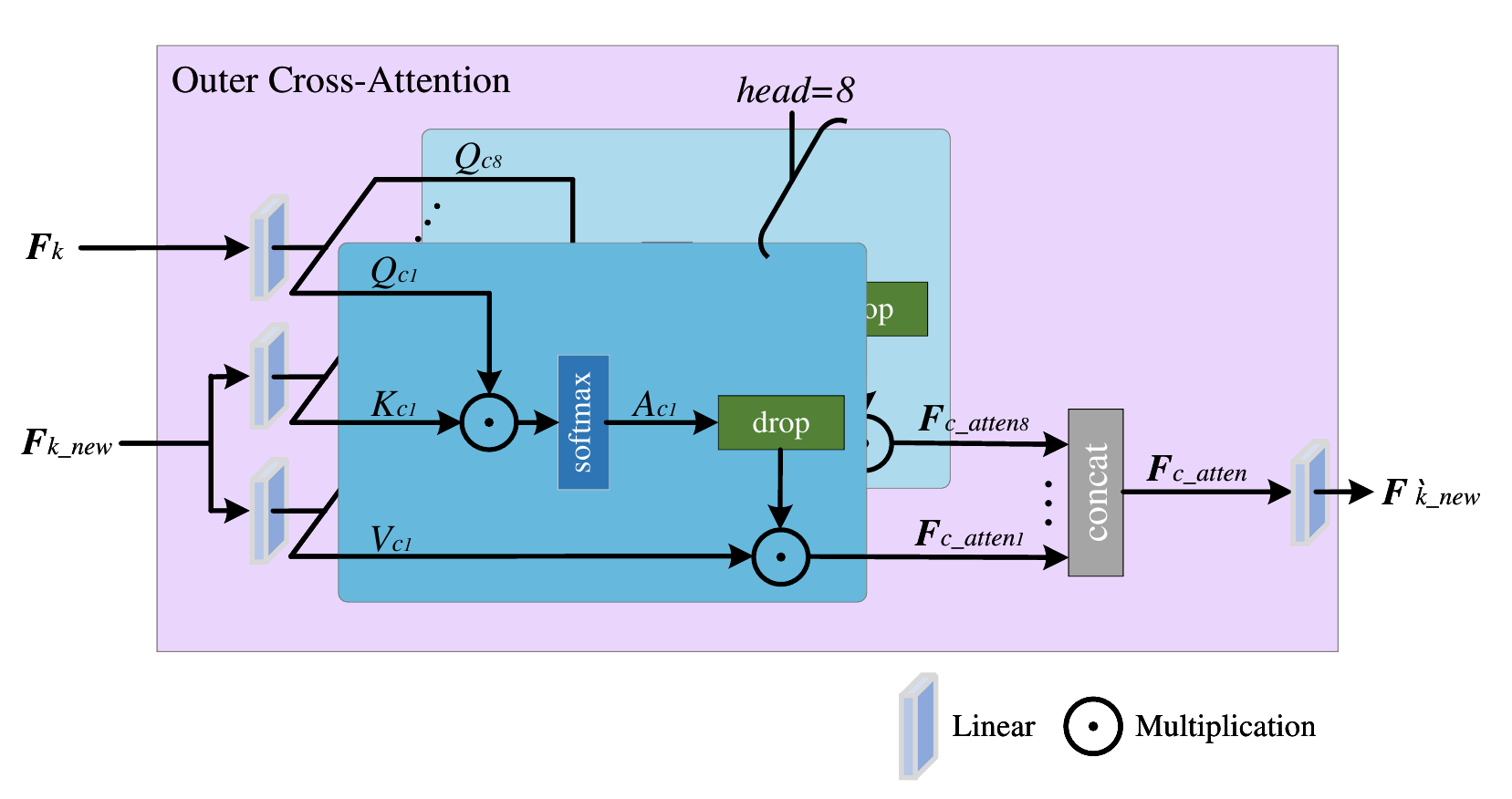}
			\caption{Outer Cross-Attention. The multi-head cross-attention method used in our model divides $Q_c$, $K_c$, and $V_c$ into eight groups, next calculates cross-attention for the eight groups respectively, and concatenates the obtained eight cross-attention encoding to obtain the multi-head cross-attention encoding $\bm{F}_{c\_atten}$, and then passes it through a linear layer to obtain the keypoint feature ${\bm{F}}'_{k\_new}$.} \label{fig:outer_atten}
		\end{figure}
		
		\textbf{Outer Cross-Attention.} After aggregating the features of local neighborhood to each corresponding keypoint, then we calculate the multi-head (head=8) cross-attention between the initial keypoint features $\bm{F}_k$ and the updated keypoint features $\bm{F}_{k\_new}$. The process for using the cross-attention mechanism to calculate the attention encoding is similar to the process by self-attention mechanism, it also uses Eq.(\ref{eq:3}) to calculate the attention weights, and finally calculate the cross-attention encoding through the Eq.(\ref{eq:4}). The difference between the two is that: the $query$, $key$, and $value$ vectors used in the self-attention calculation process are all obtained from the same feature vector through linear transformation; while in cross-attention, the $key$ and $value$ vectors come from the same feature vector, and the $query$ comes from another feature vector. The $query$ ($Q_c$), $key$ ($K_c$) and $value$ ($V_c$) matrices in the cross-attention can be expressed as:
		\begin{equation}
			\begin{cases}
				 Q_{c}=\bm{F}_{k}\cdot W_{q\_c} \\ 
				 (K_{c},V_{c})=(\bm{F}_{k\_new}\cdot W_{k\_c}\cdot \bm{F}_{k\_new}\cdot W_{v\_c})
			\end{cases} \label{eq:9}
		\end{equation}
		where $W_{q\_c}$, $W_{k\_c}$ and $W_{v\_c}$ are the shared and learnable linear transformation matrices, $Q_{c},K_{c},V_{c}\in\mathbb{R}^{M\times C_{\mathrm{out}}}$. As shown in Fig.\ref{fig:outer_atten}, the multi-head cross-attention used in this paper divides $Q_c$, $K_c$, $V_c$ into 8 sets:
		\begin{equation}
			(Q_{c},K_{c},V_{c})=\left\{(Q_{ci},K_{ci},V_{ci})\in\mathbb{R}^{M\times{\frac{c_{out}}{8}}}\vert i=1,2,\ldots,8\right\} \label{eq:10}
		\end{equation}
		where $Q_{ci}$, $K_{ci}$ and $V_{ci}$ respectively represent the $query$, $key$ and $value$ matrices of the $i$th set. The multi-head cross-attention uses these 8 sets of data to perform cross-attention calculations respectively, and the process of calculating the cross-attention weight is similar to the process of calculating the attention weight in the self-attention mechanism.
		
		To prevent overfitting, we implement random dropout on the cross-attention weights and multiply it by $V_{ci}$ to obtain the corresponding cross-attention encoding $\bm{F}_{c\_atteni}$:
		\begin{equation}
			 \bm{F}_{c\_atteni}=drop(A_{ci})\cdot V_{ci},\bm{F}_{c\_atteni}\in\mathbb{R}^{M\times\frac{C_{out}}{8}} \label{eq:11}
		\end{equation}
		where $A_{ci}\in\mathbb{R}^{M\times M}$ represents the weight of the $i$th cross-attention calculation process, and  denotes a random loss function. Without loss of generality, we set the random loss probability to 0.5 as in \cite{pct}. Then we concatenate the 8 cross-attention encodings to obtain the multi-head cross-attention encoding $\bm{F}_{c\_atten}\in\mathbb{R}^{M\times C_{out}}$:
		\begin{equation}
			\bm{F}_{c\_atten}=concat(\left\{\bm{F}_{c\_atteni}\in\mathbb{R}^{M\times{\frac{C_{out}}{8}}}\vert i=1,2,\ldots,8\right\}) \label{eq:12}
		\end{equation}
		Then we pass $\bm{F}_{c\_atten}$ through a linear layer to get the keypoint features ${\bm{F}}'_{k\_new}$:
		\begin{equation}
			{\bm{F}}'_{k\_new}=\varphi(\bm{F}_{c\_atten}),{\bm{F}}'_{k\_new}\in\mathbb{R}^{M\times C_{out}} \label{eq:13}
		\end{equation}
		where $\varphi$ denotes a linear layer.
		
		\begin{figure}
			\centering
			\includegraphics[width=0.43\textwidth]{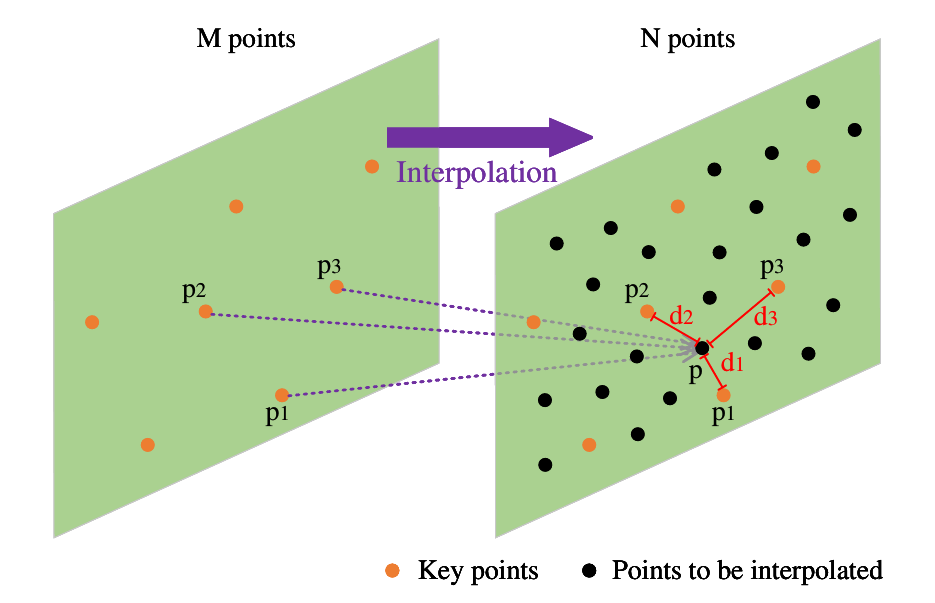}
			\caption{Three-nearest neighbor interpolation. Where $p_1$, $p_2$, $p_3$ are the three closest keypoints of the interpolated point, and $d_1$, $d_2$, $d_3$ are the Euclidean distances between these three keypoints and the interpolated point $p$, respectively. We calculate the weights according to these three distances, and then obtain the weighted summation of the three keypoints features, thus we obtain the interpolated feature of point $p$.} \label{fig:near_inter}
		\end{figure}
		
		\textbf{Upsample (Feature Interpolation).} n order to concatenate the features generated by PT Branch and V Branch, we need to upsample the keypoint feature ${\bm{F}}'_{k\_new}$ to the original point cloud through feature interpolation (mapping the feature dimension from $M$ to $N$). Here we use three-nearest neighbor interpolation to conduct the feature interpolation. The three-nearest neighbor interpolation is shown in Fig.\ref{fig:near_inter}, orange points represent M keypoints with known features, and black points denote points to be interpolated with unknown features. Assuming that the three closest keypoints of the interpolated point $p$ are $p_1$, $p_2$ and $p_3$, and the corresponding Euclidean distances between $p$ and the three keypoints are $d_1$, $d_2$, $d_3$, then the weights of the three keypoints $w_1$, $w_2$, $w_3$ can be calculated by:
		\begin{equation}
			\begin{split}
				(w_{1},w_{2},w_{3})=(d_{2}\cdot d_{3}/S&,d_{1}\cdot d_{3}/S,d_{1}\cdot d_{2}/S) \\
				S=d_{2}\cdot d_{3}+d_{1}&\cdot d_{3}+d_{1}\cdot d_{2} \label{eq:14}
			\end{split}
		\end{equation}
		Next we obtain the features $f_p$ of $p$ by the weighted summation of the three keypoints features $(f_{p_1},f_{f_{p_2}},f_{p_3})$:
		\begin{equation}
			f_{p}=f_{p_{1}}\cdot w_{1}+f_{p_{2}}\cdot w_{2}+f_{p_{3}}\cdot w_{3} \label{eq:15}
		\end{equation}
		Here, we use the three-nearest neighbor interpolation to interpolate keypoint features ${\bm{F}}'_{k\_new}\in\mathbb{R}^{M\times C_{out}}$ back to the point cloud features $F_{PT}$:
		\begin{equation}
			\bm{F}_{PT}=nearest(\bm{F}'_{k\_n e w}),\bm{F}_{PT}\in\mathbb{R}^{N\times C_{out}} \label{eq:16}
		\end{equation}
		where $nearest$ represents the three-nearest neighbor interpolation.
		
		\subsection{Point-Based Branch}
		Both V Branch and PT Branch obtain the local information of the point cloud. To make the obtained point cloud features more hierarchical, VTP block uses P Branch to obtain the global location information of the point cloud. The resulting global features, together with point-level and voxel-level local features, are used for point cloud learning.
		
		P Branch simply conducts a one-dimensional convolution layer on the input point cloud features $\bm{F}_{in}$ to obtain the global location information of the point cloud, and performs batch normalization and nonlinear activation function processing on this feature to obtain $\bm{F}_{global}$:
		\begin{equation}
			\bm{F}_{global}=\tau(\delta(\gamma(\bm{F}_{in}))),\bm{F}_{global}\in\mathbb{R}^{N\times C_{out}} \label{eq:17}
		\end{equation}
		where $\gamma$ represents a one-dimensional convolution operation, $\delta$ represents the batch normalization, and $\tau$ represents the nonlinear activation function.
		
		\subsection{Feature Combination}
		The V Branch obtains the voxel-level local feature $\bm{F}_V$, and the PT Branch obtains the point-level local feature $\bm{F}_{PT}$, as shown in the left-bottom of Fig.\ref{fig:vtpnet}, we firstly concatenate $\bm{F}_V$ and $\bm{F}_{PT}$, then use a one-dimensional convolutional neural network to map it to the output feature channel $C_{out}$, thus we get the fused local features:
		\begin{equation}
			\bm{F}_{local}=\sigma(concat(\bm{F}_{V},\bm{F}_{PT})),\bm{F}_{local}\in\mathbb{R}^{N\times C_{out}} \label{eq:18}
		\end{equation}
		where $\sigma$ denotes a one-dimensional convolutional neural network.
		
		Finally, we add the global features $\bm{F}_{global}$ generated by the P Branch and the fused local features $\bm{F}_{local}$, thus to compensate for the information loss caused by the local feature extraction:
		\begin{equation}
			\bm{F}_{out}=\bm{F}_{local}+\bm{F}_{global},\bm{F}_{out}\in\mathbb{R}^{N\times C_{out}} \label{eq:19}
		\end{equation}
			
	\section{Experiments}
	\label{sec:experiment}
	To prove the performance of VTP block, we use VTP block to construct a backbone network named VTPNet for the classification and segmentation tasks of point cloud. For the 3D part segmentation tasks, we use the widely-used ShapeNet Part dataset for experiments. For the semantic segmentation tasks of 3D scenes, we use a large indoor 3D point cloud dataset (S3DIS) provided by Stanford University for experiments. For the 3D object classification tasks, we use the ModelNet40 datasets for experiments. The whole experiment is based on the PyTorch deep learning framework, and the model is trained and tested on GeForce RTX 3080Ti GPU.
	
	\begin{figure*}
		\centering
		\includegraphics[width=1\textwidth]{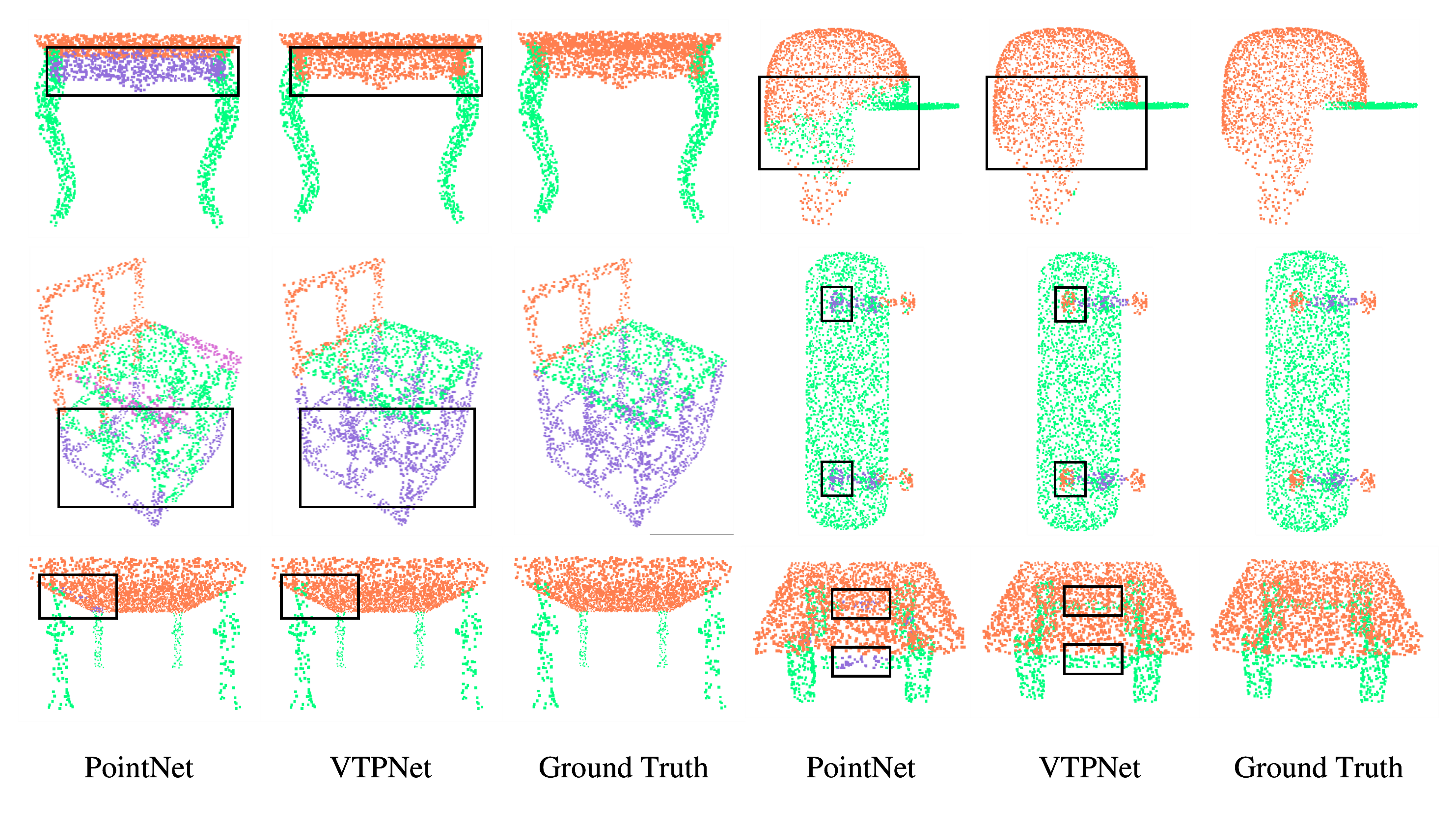}
		\caption{Visualization of part segmentation results on ShapeNet Part dataset. It contains the segmentation results of PointNet, VTPNet and the Ground Truth.} \label{fig:sapenet}
	\end{figure*}
	
	\subsection{Part Segmentation}
	\label{sec:exp_shapenet}
	\textbf{Data and metric.} We use the ShapeNet Part dataset \cite{shapenetpart} to evaluate the performance of VTPNet in the object part segmentation task. This dataset has a total of 16 categories, 16881 shapes and 50 parts, each shape contains 2 to 6 parts. In the experiment, we extract 2048 points from each shape as input, and follows the train/validation/test segmentation scheme in PointNet: 12137 shapes as the training set, 1870 shapes as the verification set, and 2874 shapes as the test set. For the evaluation metrics, we firstly obtain the IoU of a category by averaging the IoU of all shapes of a category, and then average the IoU of all shapes to obtain the mean IoU (mIoU).
	
	\textbf{Network configuration.} The network architecture for component segmentation is shown in Fig.\ref{fig:vtpnet}, and its feature extraction part consists of three VTP blocks and two MLPs with feature channels of 512 and 2048 respectively. In order to reduce the cost of memory, we use low voxel resolution in the V Branch when the VTP Block has a high number of feature channels; otherwise, we adopt high voxel resolution in V Branch. In addition, to enrich the feature representation of the point cloud, when the V Branch has a large voxel scale, the neighborhood spheres divided in the PT Branch are small and many (with a small radius and a large number); when the V Branch has a small voxel scale, the neighborhood spheres divided in the T Branch are large and small (with a large radius and a small number). In all the following experiments, the channel number is expressed as c, the voxel resolution is denoted as R, the number of sampling points is expressed as M, and the neighborhood radius is expressed as r, we use $(c, R, M, r, K)$ to represent the parameters in the VTP blocks. The parameter settings of the three VTP blocks in VTPNet for part segmentation are (64, 32, 50, 0.06, 45), (128, 16, 400, 0.03, 6) and (64, 32, 50, 0.06, 45), respectively. The experiments are trained with a batch size of 32 and 100 epochs, Adam optimizer is adopted to optimize the network model parameters, the learning rate was set to 0.0001, with an attenuation of 0.5 after every 10 epochs.
	
\begin{table*}
\renewcommand{\arraystretch}{1.3}
\setlength{\tabcolsep}{4.5pt}
\centering
\caption{Results of 3D art segmentation on ShapeNet Part dataset.}
            \label{tab:shapenet}
\scalebox{1}{
\begin{tabular}{c|c|cccccccccccccccc} 
\bottomrule
Method          & mIoU                   & \begin{tabular}[c]{@{}c@{}}air-\\plane\end{tabular} & bag                                     & cap                    & car                    & chair                  & \begin{tabular}[c]{@{}c@{}}ear-\\phone\end{tabular} & guitar                 & knife                  & lamp                   & laptop                 & \begin{tabular}[c]{@{}c@{}}motor-\\bike\end{tabular} & mug                    & pistol                 & rocket        & \begin{tabular}[c]{@{}c@{}}skate-\\board\end{tabular} & table          \\ 
\hline
PointNet\cite{pointnet}        & 83.7                   & 83.4                                                & 78.7                                    & 82.5                   & 74.9                   & 89.6                   & 73.0                                                & 91.5                   & 85.9                   & 80.8                   & 95.3                   & 65.2                                                 & 93.0                   & 81.2                   & 57.9          & 72.8                                                  & 80.6           \\
SO-Net\cite{so-net}          & 84.9                   & 82.8                                                & 77.8                                    & \textbf{88.0}          & 77.3                   & 90.6                   & 73.5                                                & 90.7                   & 83.9                   & 82.8                   & 94.8                   & 69.1                                                 & 94.2                   & 80.9                   & 53.1          & 72.9                                                  & \textbf{83.0}  \\
PointNet++\cite{pointnet++}      & 85.1                   & 82.4                                                & 79.0                                    & 87.7                   & 77.3                   & 90.8                   & 71.8                                                & 91.0                   & 85.9                   & 83.7                   & 95.3                   & 71.6                                                 & 94.1                   & 81.3                   & 58.7          & 76.4                                                  & 82.6           \\
DGCNN\cite{dgcnn}           & 85.2                   & 84.0                                                & 83.4                                    & 86.7                   & 77.8                   & 90.6                   & 74.7                                                & 91.2                   & 87.5                   & 82.8                   & 95.7                   & 66.3                                                 & 94.9                   & 81.1                   & \textbf{63.5} & 74.5                                                  & 82.6           \\
3D-GCN\cite{3d-gcn}          & 85.1                   & 83.1                                                & 84.0                                    & 86.6                   & 77.5                   & 90.3                   & 74.1                                                & 90.9                   & 86.4                   & 83.8                   & 95.6                   & 66.8                                                 & 94.8                   & 81.3                   & 59.6          & 75.7                                                  & 82.8           \\
Point-BERT\cite{yu2022point}      & 85.6                   & 84.3                                                & \textbf{\textbf{88.4}}                  & \textbf{\textbf{88.0}} & 79.8                   & 91.0                   & \textbf{\textbf{81.7}}                              & \textbf{\textbf{91.6}} & \textbf{\textbf{87.9}} & \textbf{\textbf{85.2}} & 95.6                   & 75.6                                                 & 94.7                   & \textbf{\textbf{84.3}} & 63.4          & 76.3                                                  & 81.5           \\ 
\hline
\textbf{VTPNet} & \textbf{85.8} & \textbf{84.4}                             & 87.5                & 85.6                   & \textbf{79.8}   & \textbf{91.2} & 74.1                                                & 91.2                   & 84.8                   & 84.5                   & \textbf{96.0} & \textbf{76.1}                               & \textbf{95.4} & 82.8                   & 61.3          & \textbf{77.9}                               & 82.9           \\
\bottomrule
\end{tabular}}
\end{table*}

	\textbf{Results.} The part segmentation results of our model are shown in TABLE \ref{tab:shapenet}. VTPNet achieved the best result of 85.8\% in part segmentation tasks, which is 0.2\% higher than the second-best model Point-BERT in the table. In addition, our model exhibits the best performances on airplane, car, chair, laptop, motorbike, mub, and skateboard, with the results of the motorbike increasing by approximately 0.7\% compared to Point-BERT, and the results of the skateboard increasing by approximately 2.0\% compared to PointNet++. Moreover, our model achieved the highest result of 96.0\% (Laptop) among all categories of all models. We also visualize some segmentation results in Fig.\ref{fig:sapenet}, which indicates that VTPNet performs well in part segmentation tasks.
	
	\subsection{Scene Semantic Segmentation}
	\label{sec:sep_s3dis}
	
	\begin{figure*}
		\centering
		\includegraphics[width=1\textwidth]{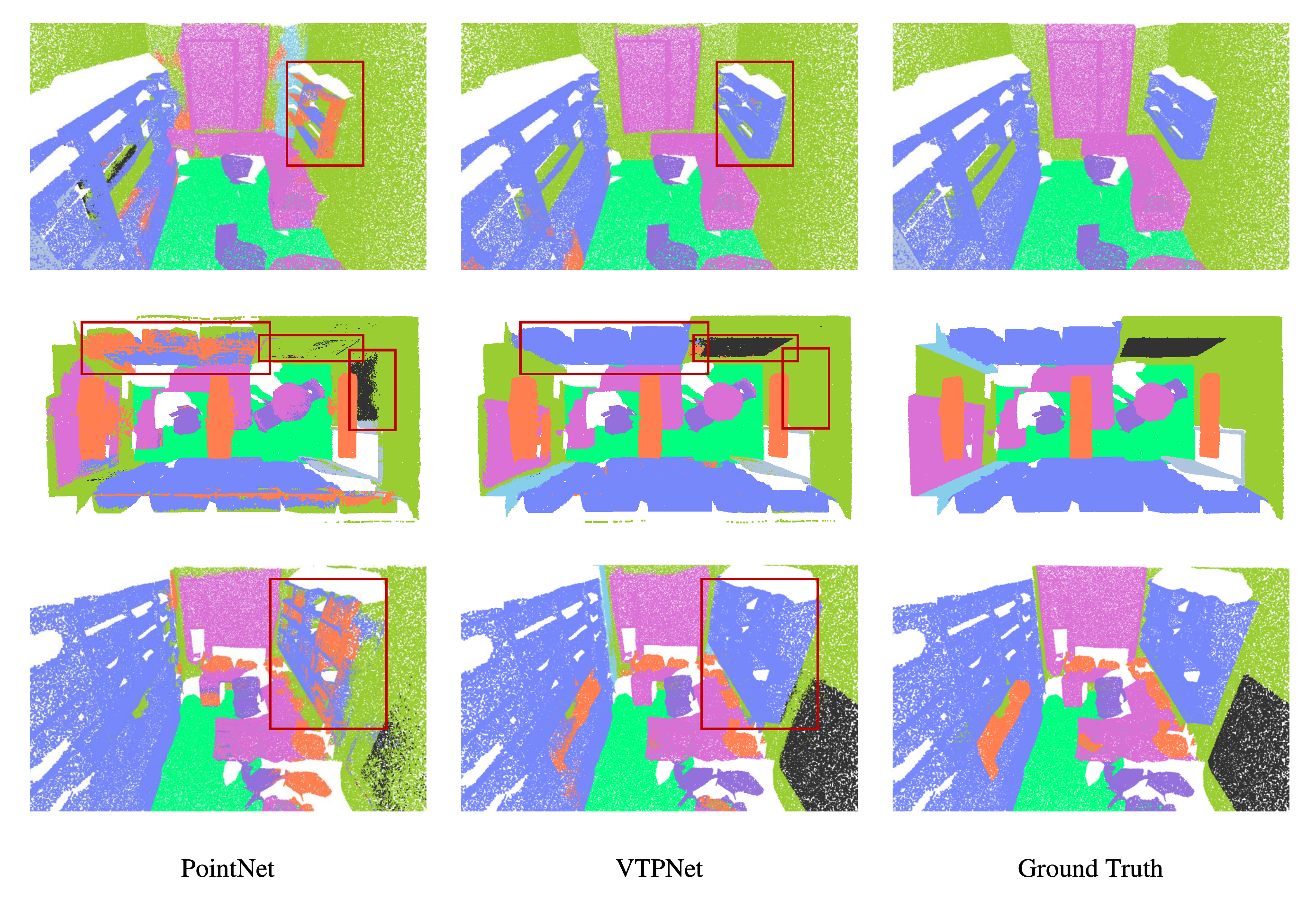}
		\caption{Visualization of semantic segmentation results on the S3DIS dataset. The first column is the segmentation result of PointNet, the second column is the segmentation result of VTPNet, and the third column is the Ground Truth.} \label{fig:s3dis}
	\end{figure*}

	\textbf{Data and metric.} To further evaluate the performance of VTP, we conduct the 3D scene semantic segmentation task on the S3DIS dataset \cite{s3dis}. S3DIS is a large-scale indoor 3D point cloud dataset provided by Stanford University, consisting of point clouds sampled from 6 areas of 3 different buildings, and contains a total of 272 rooms. Each point is annotated with one of the 12 semantic labels. This experiment follows the settings in PointNet, sampling 4096 points from a block with an area size of $1m\times 1m$ as the input of the model. We use mIoU to evaluate the performance of the model by 6-fold cross validation for 6 regions.
	
	\textbf{Network configuration.} In this section, we extend VTP to PointNet++ for 3D scene semantic segmentation, which takes the 3D coordinates, RGB values and normals of point clouds as input. In this experiment, we set the batch size to 12 with a total of 150 epochs, and use the Adam optimizer to optimize the parameters of the model. The learning rate is set to 0.0001, and we adopt CosineAnnealingLR to adjust the learning rate with T\_max of 150.
	
	\begin{table*}
 \renewcommand{\arraystretch}{1.3}
		\centering
		\caption{Results of 6-fold cross validation on the S3DIS dataset}
            \label{tab:s3dis}
        \begin{tabular}{c|c|ccccccccccccc} 
\bottomrule
Method          & miou          & celling       & floor         & wall          & beam          & column        & window        & door          & table         & chair         & sofa          & bookcase      & board         & clutter        \\ 
\hline
PointNet\cite{pointnet}        & 47.6          & 88.0          & 88.7          & 69.3          & 42.4          & 23.1          & 47.5          & 51.6          & 54.1          & 42.0          & 9.6           & 38.2          & 29.4          & 35.2           \\
SPG\cite{landrieu2018large}             & 62.1          & 89.9          & 95.1          & 76.4          & 62.8          & 47.1          & 55.3          & 68.4          & 69.2          & \textbf{73.5} & 45.9          & \textbf{63.2} & 8.7           & 52.9           \\
PointCNN\cite{li2018pointcnn}        & 65.4          & \textbf{94.8} & 97.3          & 75.8          & \textbf{63.3} & \textbf{51.7} & 58.4          & 57.2          & 71.6          & 69.1          & 39.1          & 61.2          & 52.2          & 58.6           \\
DBFA\cite{chen2022background}            & 61.6          & 94.6          & \textbf{97.7} & 77.8          & 38.5          & 38.3          & 53.3          & 67.7          & \textbf{75.2} & 66.6          & \textbf{49.8} & 49.8          & \textbf{51.4} & \textbf{60.6}  \\
JSNet++\cite{zhao2022jsnet++}         & 62.4          & 94.1          & 97.3          & 78.0          & 41.3          & 32.2          & 52.0          & \textbf{70.0} & 69.9          & 72.7          & 37.9          & 54.1          & 51.3          & 60.2           \\
SemRegionNet\cite{zhang2022semregionnet}    & 62.9          & -             & -             & -             & -             & -             & -             & -             & -             & -             & -             & -             & -             & -              \\ 
\hline
\textbf{VTPNet} & \textbf{66.3} & 94.2          & 97.4          & \textbf{81.8} & 58.3          & 43.0          & \textbf{63.7} & 67.8          & 70.5          & 72.9          & 43.3          & 58.9          & 51.2          & 59.4           \\
\bottomrule
\end{tabular}
	\end{table*}
	
	\textbf{Results.} The results of 3D indoor scene segmentation are shown in TABLE \ref{tab:s3dis}. Compared with recent models, VTPNet achieves the best result of 66.3\%, which is 0.9\% higher than PointCNN. In addition, our model has the best performance on the labels of wall and window, with the result of the wall increasing by approximately 4.9\% compared to the highest JSNet++, and the results of the window increasing by approximately 9.1\% compared to PointCNN. Part of the visualization results for S3DIS is shown in Fig.\ref{fig:s3dis}, from the boxed part in the figure, it can be observed that VTPNet is more effective in processing the details than PointNet.
	
        \subsection{3D object classification}
	\label{sec:exp_modelnet}
	\textbf{Data and metric.} ModelNet40 dataset \cite{modelnet40} consists of 40 categories of point cloud data, with a total of 12311 point cloud data that including 9843 point cloud data as the training set and 2468 point cloud data as the validation set. In this experiment, we sample 1024 points from each point cloud as input, and evaluate the performance of the model by class average accuracy (mAcc) and overall accuracy (OA).
	
	\textbf{Network configuration.} Compared to the part segmentation model, the main part of the classification model used in this experiment is only composed of 5 sequentially connected VTPs. In this experiment, our model takes the 3D coordinates and normals of the point cloud as inputs, and concatenates the point cloud features generated by 5 VTPs. The parameters $(c, R, M, r, K)$ of each VTP in the model are set to: (64, 32, 100, 0.03, 45), (128, 16, 800, 0.01, 6), (64, 32, 100, 0.03, 45), (128, 16, 800, 0.01, 6) and (64, 32, 100, 0.03, 45). Our model is trained with the batch size of 32 and 250 epochs. The Adam optimizer was used to optimize the parameters of our model, the learning rate is set to 0.001 with a decay of 0.3 after every 50 epochs.
	
	\begin{table}
        \renewcommand{\arraystretch}{1.3}
		\centering
		\caption{Results of 3D classification on the ModelNet40 dataset.}
            \label{tab:modelnet}
            \begin{tabular}{c|cc} 
\bottomrule
Method          & mAcc(\%)      & OA(\%)         \\ 
\hline
PointNet\cite{pointnet}        & 86.0          & 89.2           \\
PointNet++\cite{pointnet++}      & -             & 91.9           \\
DGCNN\cite{dgcnn}           & \textbf{90.2} & 92.9           \\
3D-GCN\cite{3d-gcn}          & -             & 92.1           \\
DRNet\cite{qiu2021dense}           & -             & 93.1           \\
PCT\cite{pct}             & -             & \textbf{93.2}  \\ 
\hline
\textbf{VTPNet} & 90.1          & \textbf{93.2}  \\
\bottomrule
\end{tabular}
	\end{table}
	
	\textbf{Results.} The results of this experiment are shown in TABLE \ref{tab:modelnet}. VTPNet achieves comparable results to other methods and has the highest overall accuracy of 93.2\%, which is 4.1\% higher than PointNet. Compared with the Transformer-based model PCT, VTPNet owns a simpler model structure while achieving the same overall accuracy.

	\subsection{Ablation Study}
	\label{sec:exp_ablation}
	To investigate the effectiveness of different structures in VTP, we conduct extensive ablation studies on the ShapeNet Part dataset.

	\begin{table}
 \renewcommand{\arraystretch}{1.3}
		\centering
		\caption{The impact of voxel scale and neighborhood sphere scale. Large and small voxel scales correspond to voxel resolutions of 16 and 32, respectively, while large and small neighborhood sphere scales indicate a neighborhood sphere radius of 0.03m and 0.01m, respectively.}
		\label{tab:ablation_3}
		\begin{tabular}{c|cc|c} 
			\bottomrule
			Number             & Voxel Scale & Neighborhood Sphere Scale & mIoU(\%)                                  \\ 
			\hline
			\multirow{2}{*}{1} & large       & large                              & \multirow{2}{*}{85.63}                    \\ 
			\cline{2-3}
			& small       & large                              &                                           \\ 
			\hline\hline
			\multirow{2}{*}{2} & large       & small                              & \multirow{2}{*}{85.77}                    \\ 
			\cline{2-3}
			& small       & small                              &                                           \\ 
			\hline\hline
			\multirow{2}{*}{3} & large       & large                              & \multirow{2}{*}{85.66}                    \\ 
			\cline{2-3}
			& small       & small                              &                                           \\ 
			\hline\hline
			\multirow{2}{*}{4} & large       & small                              & \multirow{2}{*}{\textbf{\textbf{85.82}}}  \\ 
			\cline{2-3}
			& small       & large                              &                                           \\
			\bottomrule
		\end{tabular}
	\end{table}
	
	\textbf{The impact of voxel scale and neighborhood sphere scale.} The V branch and PT branch in VTP are used to extract voxel-level and point-level local features, respectively. The V branch needs to divide the point cloud into regular voxel grids, while the PT branch needs to divide the point cloud into partially overlapping neighborhood spheres. We consider the relationship between the voxel scale and the neighborhood sphere scale used in these two branches and conduct experiments on four cases, where large and small voxel scales correspond to voxel resolutions of 16 and 32, respectively, and large and small neighborhood sphere scales represent neighborhood sphere radius of 0.03m and 0.01m, respectively. The experimental results are shown in TABLE \ref{tab:ablation_3}. From the results, we can observe that setting of large voxel scale \& small neighborhood sphere scale or small voxel scale \& large neighborhood sphere scale yields the best performance of 85.82\%, which is also the setting used in our model.
	
	\begin{table}
 \renewcommand{\arraystretch}{1.3}
		\centering
		\caption{The impact of features used for Inner self-attention computation on the model.}
		\label{tab:ablation_4}
\begin{tabular}{c|c}
  \bottomrule
    Feature & mIoU(\%)                \\
    \hline
    $(\bm{F}_{neighbor_i})$     & 85.70                   \\
    $(\bm{F}_{neighbor_i}-f_{k_i})$     & 85.53 \\
    $(\bm{F}_{neighbor_i}-f_{k_i},\bm{F}_{neighbor_i})$     & 85.69 \\
	$(\bm{F}_{neighbor_i}-f_{k_i},f_{k_i})$     & \textbf{85.82} \\
 $(\bm{F}_{neighbor_i}-f_{k_i},f_{k_i},\bm{F}_{neighbor_i})$     & 85.76 \\
    \bottomrule
\end{tabular}

	\end{table}
	
	\textbf{The impact of features used for Inner self-attention computation on the model.} To choose the most suitable feature combination model for self-attention calculation in the neighborhood, we set up 5 sets of feature combination models, and the experimental results are shown in TABLE \ref{tab:ablation_4}, where $f_{k_i}$ denotes the $i$th keypoint feature and $\bm{F}_{neighbor_i}$ denotes the neighborhood point feature in the neighborhood of the $i$th keypoint. The result of only using neighborhood point features is 85.70\%, while concatenating the keypoint features and the difference features between neighborhood points and keypoints obtain the highest result of 85.82\%. This is also the feature combination model in VTP for the calculation of self-attention. Although the concatenation of the difference features between neighborhood points and key points, keypoint features, and neighborhood point features obtains a good result of 85.76\%, there is still a gap with the best result obtained by the combination used in our model.
	
	\begin{table}
 \renewcommand{\arraystretch}{1.3}
		\centering
		\caption{The impact of feature aggregation type in Middle Aggregation Layer. A: max pooling, B: average pooling, C: difference between features obtained by max pooling and average pooling, D: concatenation of features obtained by max pooling and average pooling.}
		\label{tab:ablation_5}
		\begin{tabular}{c|c}
  \bottomrule
			Aggregation Type          & mIoU(\%)                \\
   \hline
			A  &85.62                  \\
			B  &85.72\\
			C  &85.61\\
			D  &\textbf{85.82} \\
			\bottomrule
		\end{tabular}
	\end{table}
	
	\textbf{The impact of feature aggregation type in Middle Aggregation Layer.} The Middle Aggregation Layer aggregates the point features within the neighborhood sphere to keypoints. To find a suitable aggregation manner, we carry out experiments on four aggregation manners, and the experimental results are shown in TABLE \ref{tab:ablation_5}. Only using the max pooling aggregation features, the accuracy can reach 85.62\%; using only the average pooling, the accuracy can reach 85.72\%; the features obtained by the subtraction of the max pooling and the average pooling obtain the accuracy of 85.61\%; the features concatenated by the max pooling and the average pooling obtain the highest accuracy of 85.82\%.
	
	\section{Conclusion}
	In this paper, we propose VTP for 3D point cloud feature learning and construct VTPNet network architecture for different 3D point cloud tasks. The V, PT and P branches in VTP capture the voxel-level local features, point-level local features and global location features of the point cloud respectively, which provide point cloud features at different levels and scales for point cloud analysis. In addition, the setting of V branch voxel scale and PT branch neighborhood sphere scale in this paper enables VTP to capture richer local features, which promotes the applicability of the network model under different datasets. The experimental results show that the VTPNet network constructed in this paper can effectively obtain the local geometric structure information of point clouds.
	
	It is hoped that the future work can be improved from the following three aspects: (1) improve the ability of VTPNet to extract the global features of point clouds, and design a more effective global feature extraction branch to replace the P branch in VTP; (2) improve the overall structure of VTPNet, such as Encoder-Decoder structure; (3) extend VTPNet to other 3D point cloud analysis tasks, such as object detection and point cloud completion.
		
	\bibliographystyle{IEEEtran}
	\bibliography{mybibfile}

	\end{document}